\documentclass[conference]{IEEEtran}
\IEEEoverridecommandlockouts
\pdfoutput=1
\usepackage{amsmath,amssymb,amsfonts}
\usepackage{cite}
\usepackage{algorithm}
\usepackage{algorithmic}

\usepackage{subfig}
\usepackage{mwe}
\usepackage{graphicx}
\usepackage{textcomp}
\usepackage{xcolor}
\usepackage{amsmath}
\usepackage{svg}
\usepackage{hyperref}
\usepackage[numbers,sort&compress]{natbib}
\usepackage{threeparttable}
\usepackage{multicol}
\usepackage{multirow}
\usepackage{booktabs}
\usepackage{geometry}
\usepackage{amsmath}

\svgsetup{
    inkscapepath=i/svg-inkscape/
}
\svgpath{{svg/}}
\def\BibTeX{{\rm B\kern-.05em{\sc i\kern-.025em b}\kern-.08em
    T\kern-.1667em\lower.7ex\hbox{E}\kern-.125emX}}
\begin{document}
\title{HeR-DRL:Heterogeneous Relational Deep Reinforcement Learning for Decentralized Multi-Robot Crowd Navigation\\

\thanks{
	$^1$X. Zhou and S. Piao are with the Faculty of Computing, Harbin Institute of Technology, Harbin, 150001 China. E-mail: zhouxy@stu.hit.edu.cn, piaosh@hit.edu.cn.}
\thanks{
	$^2$W. Chi, L. Chen and W. Li are with Soochow University, Soochow, 215031, China. E-mail: wzchi@suda.edu.cn, chenliguo@suda.edu.cn, li7713@163.com}

\thanks{$^{*}$S. Piao is the Corresponding author.}

\thanks{This work has been submitted to the IEEE for possible publication. Copyright may be transferred without notice, after which this version may no longer be accessible.}
}

\author{\IEEEauthorblockN{Xinyu Zhou$^1$, Songhao Piao$^{1*}$, Wenzheng Chi$^2$, Liguo Chen$^2$, and Wei Li$^2$}
}
\maketitle

\begin{abstract}
Crowd navigation has received significant research attention in recent years, especially DRL-based methods. While single-robot crowd scenarios have dominated research, they offer limited applicability to real-world complexities. The  heterogeneity of interaction among multiple agent categories, like in decentralized multi-robot pedestrian scenarios, are frequently disregarded. This "interaction blind spot" hinders generalizability and restricts progress towards robust navigation algorithms. In this paper, we propose a heterogeneous relational deep reinforcement learning(HeR-DRL), based on  customised heterogeneous GNN, in order to improve navigation strategies in decentralized multi-robot crowd navigation. Firstly, we devised a method for constructing robot-crowd heterogenous relation graph that effectively simulates the heterogeneous pair-wise interaction relationships. We proposed a new heterogeneous graph neural network for transferring and aggregating the heterogeneous state information. Finally, we incorporate the encoded information into deep reinforcement learning to explore the optimal policy. HeR-DRL are rigorously evaluated through comparing it to state-of-the-art algorithms in both single-robot and multi-robot circle crowssing scenario. The experimental results demonstrate that HeR-DRL surpasses the state-of-the-art approaches in overall performance, particularly excelling in safety and comfort metrics. This underscores the significance of interaction heterogeneity for crowd navigation. The source code will be publicly released in https://github.com/Zhouxy-Debugging-Den/HeR-DRL.
\end{abstract}
\section{Introduction}
Recently, to fulfill essential daily service needs for individuals, the frontiers of mobile robot system research has transitioned from unmanned factories to dynamic environments coexisting with humans, such as offices\cite{huber2022avoiding}, hospitals\cite{kodur2023patient}, canteens\cite{fan2019getting} and other public places. Robot navigation is evolving, aiming for superior safety, efficiency, and user comfort. This comfort focus prioritizes minimizing disruptions and managing stress\cite{kruse2013human}. Nevertheless, navigating crowded, unpredictable environments remains challenging due to complex pedestrian movements and the difficulty of modeling comfort.

The research on navigation in pedestrian-rich scenarios is frequently called social navigation or crowd navigation. In this paper, we will refer to it as crowd navigation  for convenience. Throughout the past 30 years, the solutions to crowd navigation have been generally classified into three primary categories: reaction-based, trajectory-based, and DRL-based methods. Reaction-based methods automate immediate actions against obstacles based on predefined rules, i.e. RVO\cite{van2008reciprocal}, ORCA\cite{van2011reciprocal}, SFM\cite{helbing1995social}, IGP\cite{trautman2010unfreezing}. However, its decision-making process primarily hinges on the current state, highly likely triggering the short-sighted behaviour. Moreover, there's a risk of ``reciprocal dance" on account of neglecting pedestrian reactions\cite{feurtey2000simulating}. Trajectory-based methods devise the feasible planning accordingly after predicting the intended trajectories of other agents \cite{aoude2013probabilistically}, \cite{kretzschmar2016socially}, \cite{trautman2013robot}, \cite{chen2020relational}, \cite{cao2019dynamic}. Unfortunately, the computational expense of online prediction and path search in a vast state space can be significant\cite{driggs2017integrating}. In addition, predicted trajectories enlarge the spatially infeasible region, prone to result in overly conservative robot movement\cite{driggs2018robust} and the more serious ``freezing robot problem"\cite{trautman2010unfreezing}.
\begin{figure}[t]
    \includegraphics[scale=0.32]{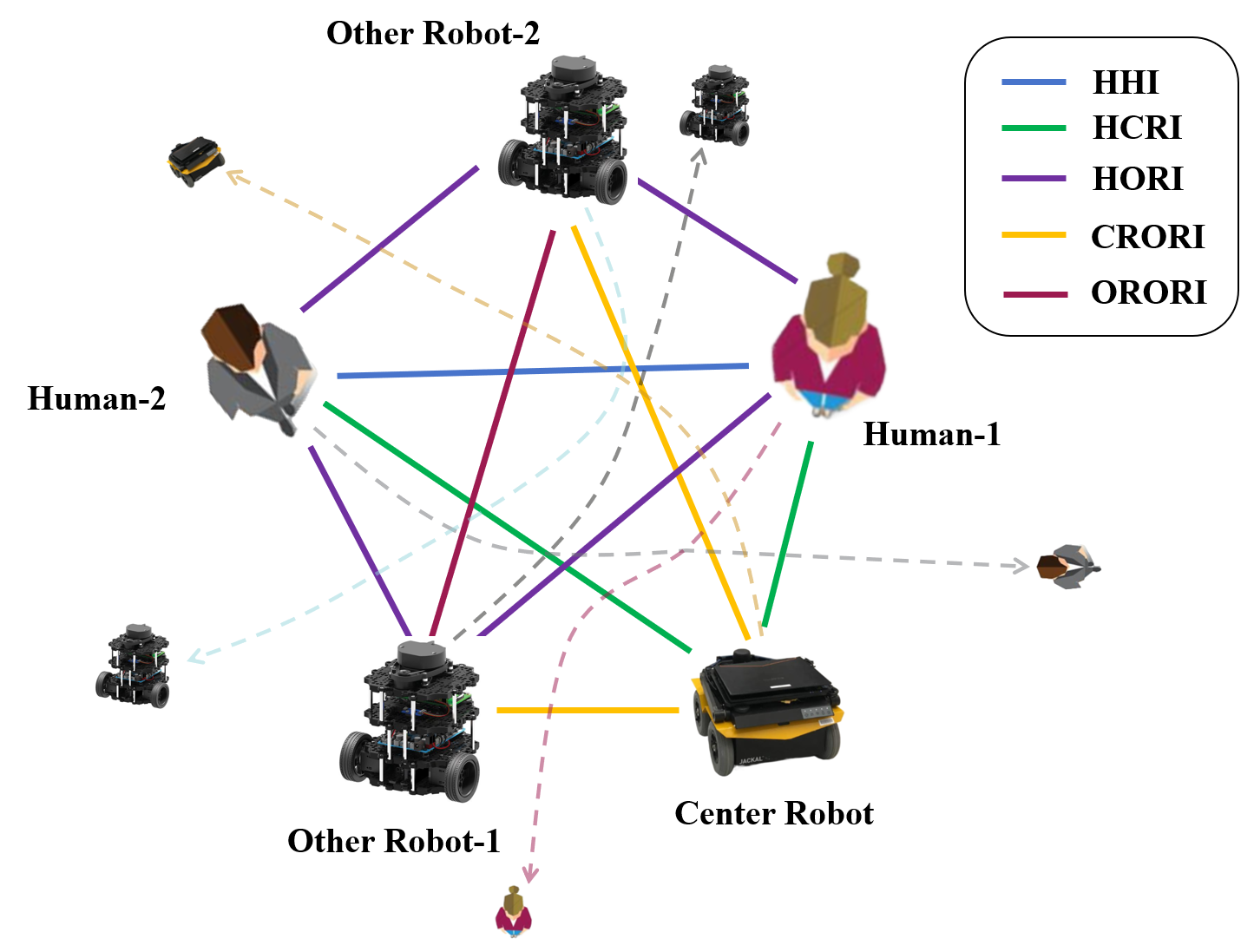}
    \centering
    \caption{Heterogeneous interaction relationships in decentralized multi-robot crowd navigation. The center robot is controlled by trained policy, while the other robots controlled by other unknown policies. Consequently, there are five heterogeneous pair-wise interactions: HHI refers to the interaction specifically occurring among human, HCRI refers to the interaction between human and the center robot, HORI refers to the interaction between human and other robots, CRORI refers to the interaction between the center robot and other robots, and ORORI refers to the interaction specifically occurring among other robot.}
    \label{fig:my_label}
\end{figure}

Deep Reinforcement Learning (DRL) offers an alternative method to implicitly merge interactive prediction and planning\cite{chen2017decentralized}, \cite{everett2018motion}, \cite{chen2019crowd}, \cite{zhang2021relational}, \cite{chen2020relational}, \cite{yang2023st}. This effectively shifts the burdensome online computation to an offline training procedure. Crowd navigation exhibit high dynamism and complexity, making the effective extraction of interaction information a challenging endeavor. Numerous explorations have structured deep learning approaches for encoding interaction information, through employing various models, such as LSTM (Long Short-Term Memory)\cite{everett2018motion}, Self-Attention Mechanisms\cite{chen2019crowd}, GCN (Graph Convolutional Networks)\cite{zhang2021relational}, \cite{chen2020relational}, Transformers\cite{yang2023st}, and more.

However, these studies often assume that the relations among agents are homogenous, disregarding the inherent heterogeneity originating from the discrepancy in the interaction relationships. This will lead to under-constructed associations between agents in crowd navigation, and social performance will inevitably be limited. The more complex the scenario, the more pronounced the impact of heterogeneity. In particular, a few recent studies\cite{wang2023multi}, \cite{chandra2023decentralized} have focused on multi-robot crowd navigation, mainly in view of the low efficiency of single-robot crowd navigation. Regrettably, these studies primarily tackle the issue of multi-agent cooperative control without concentrating on the heterogeneity of complex interactions. Therefore, this paper introduces the pioneering notions of heterogeneous interaction relationships in decentralized multi-robot crowd navigation, as depicted in Fig. 1. Building upon this viewpoint, a heterogeneous GNN-based approach called HeR-DRL is proposed. The primary contributions of this letter are outlined as follows.

\begin{enumerate}
\item We introduce a innovative heterogeneous relational deep reinforcement learning(HeR-DRL) to represent the interactive states. HeR-DRL attains safe and comfort performances in the multi-robot crowded scenarios. 
\item We constructed a novel form of heterogeneous graph neural network to extract diverse pairwise spatial interactions. New state information for each category  is configured to enhance the network in effectively encodding the heterogeneity of interactions.
\item A decentralised multi-robot 2D circle crossing simulation is innovatively designed. Experimental results in a variety of configurations demonstrate that our approach outperforms previous state-of-the-art methods in safety and comfort, and  presents exceptionally superior generalization.
\end{enumerate}
\section{RELATED WORKS}
In 2017, Yufan Chen et al.\cite{chen2017decentralized} introduced the CADRL method, pioneering a DRL-based algorithm to illustrate interactions between pairs of humans and robots. However, its use of a fixed-dimension input in a typical feed-forward neural network led to decreased scene adaptability due to the variable number of agents in the scenario. Subsequently, in 2018, Michael Everett et al.\cite{everett2018motion} proposed the GA3C-CADRL algorithm, addressing the CADRL's fixed dimensionality issue by employing LSTM to encode spatial relationships of HRI(Human-Robot Interaction). Nevertheless, this approach overlooked the presence of HHI(Human-Human Interaction) in the scenario. Consequently, Changan Chen et al.\cite{chen2019crowd} revisited the pairwise interaction and proposed a new approach called SARL, utilizing self-attention to jointly model HRI and HHI. Although this method only provided a locally coarse-grained characterization of HHI, it exhibited promising performance. Building on prior work, Weixian Shi et al.\cite{shi2022enhanced} propose the ESA by amalgamating LSTM and self-attention mechanisms, leveraging the spatial graph as a parallel branch of the modified attention graph. However, most of the aforementioned methods primarily concentrate on modeling HRI spatial interactions using various neural networks, with comparatively lesser emphasis on HHI spatial interactions.

As research progressed, successive studies began adopting a relational graph perspective to address this challenge. Changan Chen et al.\cite{chen2020relational} , for instance, introduces a relational graph to understand the relationship between HHI and HRI, employing a bi-layered GCN for computing interactions among agents, showcasing superior performance compared to SARL. Yuying Chen et al.\cite{chen2020robot} utilizes a bi-layered GCN to train an attention network, predicting human attention. Xueyou Zhang et al.\cite{zhang2021relational} captures spatial relationships of HRI and HHI using GCNs. Zhiqian Zhou et al.\cite{zhou2022robot} explores integrating social attention mechanisms to more effectively extract spatial relationships between HHI and HRI using GAT(Graph Attention Network). While the mentioned approach encodes HHI and HRI spatial interactions, it doesn't explicitly differentiate between the relationship characteristics inherent in HRI and HHI.

Subsequent research has emphasized that agent associations aren't limited to the spatial domain, changes in the temporal domain can also significantly impact interactions. Shuijing Liu et al.\cite{liu2021decentralized} proposed a new method called DSRNN, which utilises capturing spatio-temporal interactions between robots and multiple pedestrians. While it integrates spatio-temporal features of human-robot interaction to some degree, there's limited description of spatio-temporal features of HHI. The later proposed a new approach incorporates intent information and employs temporal edges to establish a comprehensive social environment graph, aiming to portray potential associations of multi-robot systems\cite{liu2023intention}. Despite integrating spatio-temporal features of HHI and HRI, compared to DSRNN, it doesn't adequately model potential spatio-temporal associations. Additionally, recent approaches employing transformers for feature representation, such as ST$^2$\cite{yang2023st}, exhibit remarkable performance in encoding spatio-temporal features and associations of HHI and HRI. Nonetheless, these approaches haven't yet explored the heterogeneity present in the multi-agent crowd navigation.
 \begin{figure*}[t]
	\centering
	\includegraphics[width=0.9\textwidth]{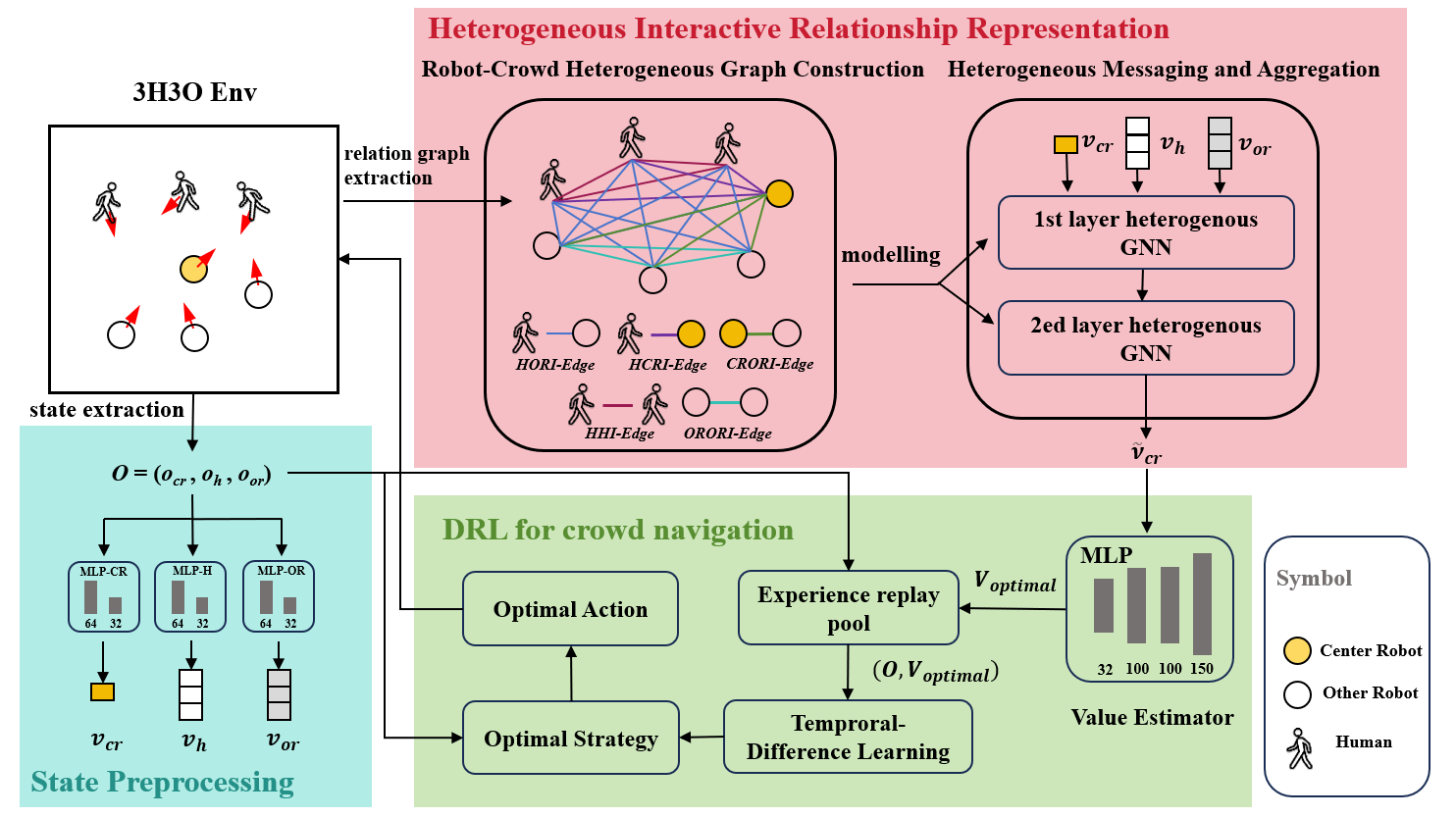}
	\caption{The overall framework of our HeR-DRL. Here is an illustration of a multi-robot crowd scenario with a 3H3O configuration.} 
	\label{fig:my_label}
\end{figure*}
Kai Zhu et al.\cite{zhu2022collision} introduced DH-SARL, an approach utilizing OBC(Orientated Bounding Capsule) and velocity-related collision risk function to capture the shape and velocity heterogeneity of agents. Aided by reward shaping, it optimizes the success rate and security of SARL. However, this method disregards interaction heterogeneity, where various types of interactions affect each other differently. Consequently, to fill the research gap, this paper concentrates on investigating the spatial heterogeneity within interaction relations, specifically in decentralized multi-robot crowd navigation. 
\section{METHOD}
Crowd navigation can be conceptualized as a POMDP (Partially Observable Markov Decision Process). The concrete workflow of the method is illustrated in Fig. 2. Step one involves extracting agents state information from the scenario and embedding it into the same-dimensional vector. In the second step, a robot-crowd heterogeneous relation graph will be constructed to transfer and  aggregate heterogeneous message using two-layer heterogenous graph neural network(heterogenous GNN). Ultimately, center robot node incorporated with interaction relations of the neighbour agents, are utilized as input for the value network to acquire optimal strategy. 
\subsection{State Information Preprocessing}
In this section, the significant state information within a multi-robot crowd navigation are initially parameterised. These details are preprocessed as node information for the subsequent construction of the robot-crowd heterogeneous relation graph.
\subsubsection{State Parameterisation}
In previous generic state parameterization definitions\cite{chen2019crowd}, \cite{yang2023st}, \cite{chen2020relational}, the state space comprises both observable and unobservable states of all agents. The observable state includes the velocity \textbf{v} = [$v_x$, $v_y$], position \textbf{p} = [$p_x$, $p_y$] and and the radius $r$ of the agent itself. Meanwhile, the unobservable state comprises the target position \textbf{p$_g$} = [$g_x$, $g_y$], preferred velocity $v_{pref}$ and the direction angle $\psi$. Compared to single-robot crowd navigation, decentralised multi-robot crowd navigation introduces an additional category of agents, namely "other robots", as shown in Fig. 1. To make a distinction, the robot our policy control are designated as the "center robot". The center robot is fully observable, whereas the humans and the other robots are partially observable. We use $o_{cr}$ , $o_{h}=[o_{h_1},o_{h_2},...,o_{h_n}]$ ,  and $o_{or}=[o_{or_1},o_{or_2},...,o_{or_m}]$ to represent the state of the center robot ,the humans and other robots respectively. 
\begin{align}
	o_{cr} &= [d_g,v_{pref},\theta,\textit{$r$},v_x,v_y], \notag \\
	o_{h_i}&=[p^{h_i}_x,p^{h_i}_y,v^{h_i}_x,v^{h_i}_y,r^{h_i},d^{h_i}_a,r^{h_i}+r,c^{h_i}], \\
	o_{or_j}&=[p^{{or}_j}_x,p^{{or}_j}_y,v^{{or}_j}_x,v^{{or}_j}_y,r^{{or}_j},d^{{or}_j}_a,r^{{or}_j}+r,c^{{or}_j}], \notag
\end{align}
where $o^i_h$ is the observable state of the $i$-th human, $o^j_{or}$ is the observable state of the $j$-th other robot, $d_g = \Vert \textbf{p} - \textbf{p}_g \Vert_2$ is the center robot's distance to the goal, $d_a^* = \Vert \textbf{p} - \textbf{p}_* \Vert_2$ is the robot's distance to the agent $*$. Notably, compared to previous studies\cite{chen2019crowd}, \cite{yang2023st}, this paper proposes a new state parameter, the agent category $c^*$ which can be received by object detection\cite{redmon2016you} in the real world. 1 if human, 0 if other robot. This parameter is established with the attention to capture the category heterogeneity of all agents except center robot, thus boosting the extraction of information about the heterogeneous relations.
\subsubsection{State Embedding}
In this section, $o_{cr}$, $o_{h}$, $o_{or}$ are embedded into a fixed-length embedding vector $v_{cr}$, $v_{h}$, $v_{or}$ separately, employed as the node of the subsequent heterogenous relation graph.
\begin{align}
	v_{cr} &= f_{cr}(o_{cr},W^{embedding}_{cr}), \notag \\
	v_{h}  &= f_{h_i}(o_{h},W^{embedding}_{h}),\\
	v_{or} &= f_{or_i}(o_{or},W^{embedding}_{or}),\notag
\end{align}
where $f_{cr}$, $f_{h}$ and $f_{or}$ are the distinct 2-layer MLP with ReLU activations, the weights are denoted by $W^{embedding}_{cr}$, $W^{embedding}_{h}$ and $W^{embedding}_{or}$ distinctly.
\subsection{Heterogeneous Interactive Relationship Representation }
This part focuses on capturing complex interactions among agents based on relational graph learning\cite{chen2020relational}. Firstly, We introduce a method for constructing the robot-crowd heterogeneous graph, facilitating the extraction of diverse categories of interactions. In the second step, we will innovatively employ a heterogeneous graph neural network approach to characterisation of robot-crowd heterogeneous graph for information transfer among different nodes.

\subsubsection{Robot-Crowd Heterogeneous Graph Construction}
In previous graph-based studies\cite{chen2020relational}, \cite{chen2020robot}, \cite{zhang2021relational}, graphs were typically constructed under the assumption of homogeneity. While these graphs encoded both HRI and HHI, they failed to differentiate between these two types of interactions. Consequently, Limited information encoding hampered the network's effectiveness, potentially hindering its performance in decentralized multi-robot crowd navigation.
Real-world graphs often exhibit a diverse array of nodes and edges, a concept widely recognized as a heterogeneous information network(HIN)\cite{shi2016survey}. In this context, we construct the robot-crowd heterogeneous graph, aiming to comprehensively capture the intricacies inherent in multi-robot crowd navigation scenarios.

In this letter, the robot-crowd heterogeneous graph, denoted as the $\mathcal{G}_{general} = (\mathcal{A},\mathcal{L})$, consists of an agents set $\mathcal{A}$ and a link set $\mathcal{L}$. The robot-crowd heterogeneous graph is also associated with a node type mapping function $\varsigma : \mathcal{A} \rightarrow \mathcal{V}$ and a link type mapping function $\tau : \mathcal{L} \rightarrow \mathcal{E}$. where $\mathcal{V}$ is the set of types of embedding features and $\mathcal{E}$ is the set of types of edges. This mapping function $\varsigma$ is actually the state embedding part mentioned above, so the $\mathcal{V} = \left \{ v_{cr}, v_{h}, v_{or} \right \} $. As illustrated in Fig. 1, this letter proposes five distinct types of pair-wise interactions. when constructing the robot-crowd heterogeneous graph, a total of five different edges exist, as shown in Fig. 2. so the $\mathcal{E} = \left \{ \mathcal{E}_{HORI}, \mathcal{E}_{HCRI}, \mathcal{E}_{CRORI}, \mathcal{E}_{HHI}, \mathcal{E}_{ORORI} \right \} $.

Next, We break down the robot-crowd interaction network(composed of all pairwise interactions) into five smaller networks, each focusing on a specific type of interaction. These five homogeneous subgraphs actually are represented as the five relations in Fig. 3. For ease of expression, we define the subgraph $\mathcal{G}_{CRHI}$ to be the relation between the center robot and the humans, the subgraph $\mathcal{G}_{ORHI}$ to be the relation between the other robots and the humans, the subgraph $\mathcal{G}_{CRORI}$ to be the relation between the center robot and the other robots, the subgraph $\mathcal{G}_{HHI}$ to be the relation among the humans themselves, the subgraph $\mathcal{G}_{ORORI}$ to be the relation among the other robots themselves. The specific formula is as follows.
\begin{align}
	\mathcal{G}_{CRHI} &= (\mathcal{V}_{(v_{cr},v_h)},\mathcal{E}_{HCRI}), \notag \\
	\mathcal{G}_{ORHI} &= (\mathcal{V}_{(v_{or},v_h)},\mathcal{E}_{HORI}), \notag \\
	\mathcal{G}_{ORCRI} &= (\mathcal{V}_{(v_{or},v_{cr})},\mathcal{E}_{CRORI}), \\
	\mathcal{G}_{HHI} &= (\mathcal{V}_{(v_h,v_h)},\mathcal{E}_{HHI}), \notag \\
	\mathcal{G}_{ORORI} &= (\mathcal{V}_{(v_{or},v_{or})},\mathcal{E}_{ORORI}), \notag 
\end{align}
In the above equation, $\mathcal{V}_{(*,*)}$ refers to the node category of each pair of interaction relations.
\begin{figure}[t]
	\includegraphics[scale=0.38]{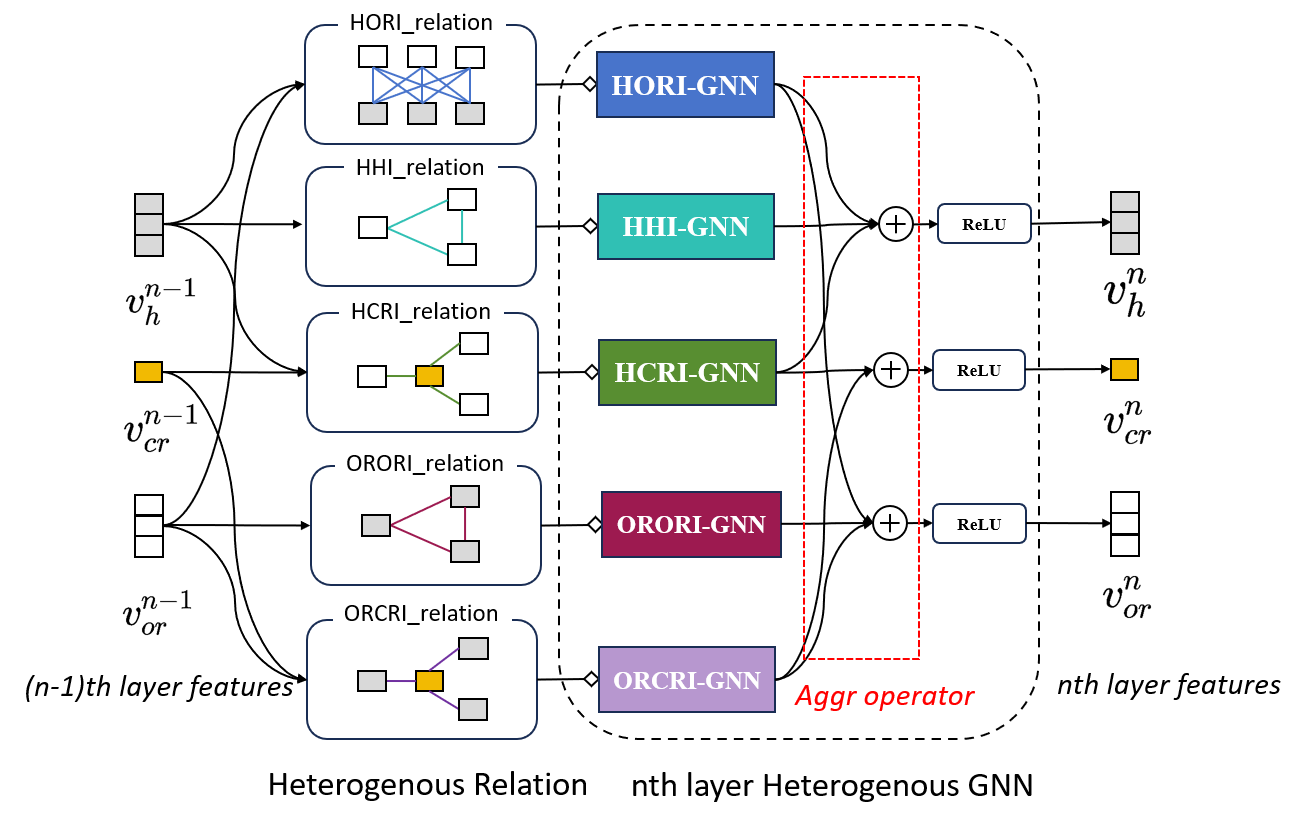}
	\centering
	\caption{Diagram of the nth-layer heterogeneous GNN with a 3H3O
		configuration.}
	\label{fig:my_label}
\end{figure}
\begin{align}
	\tilde{v}_{cr}=F^2_{Hetero}(F^1_{Hetero}((v_{cr},v_{h},v_{or}))),
\end{align}
where $\tilde{v}_{cr}$ is the center robot node-level feature after the processing of heterogeneous messaging and aggregation.
\subsubsection{Heterogeneous Messaging and Aggregation}
In previous studies, many models like GCN\cite{chen2020relational}, \cite{chen2020robot} and GAT\cite{zhou2022robot} were commonly employed to handle data aggregation in relational graphs. However, these graph neural networks are limited in their ability to accommodate diverse node and edge types, making them suitable only for homogeneous graphs.In this section we propose a new heterogeneous GNN for processing the robot-crowd heterogeneous graph constructed in the previous section.

First, inspired by \cite{hamilton2017representation}, we apply the following graphical neural network (GNN) operator to each of the five different types of homogeneous relation subgraphs. The rationale behind GNN lies in the concept that graphs can be viewed as frameworks for "information transfer" algorithms between nodes, aligning more with the core principles of this letter's heterogeneous interaction information setting, rather than  GCN or GAT which mere aggregations of information from neighboring nodes. The following outlines the computation process of a layer of our GNN operators.
\begin{align}
	h^{(n)}(v) =  h^{(n-1)}(v)\cdot W^{(n)}_1+\sum_{w \in N(v)}h^{(n-1)}(w)\cdot W^{(n)}_2 ,
\end{align}
where $h^{(n-1)}(v)$ is the $v$ node-level feature of layer $n-1$, $h^{(n-1)}(w)$ is the $v$ 's neighbour node-level feature at layer $n-1$. $W^n_1$, $W^n_2$ are layer-specific trainable weight matrix when passed from layer $n-1$ to layer $n$ respectively. 

We employ the mentioned GNN operators for each heterogeneous subgraph, forming ORCRI-GNN, HCRI-GNN, ORORI-GNN, HHI-GNN, ORHI-GNN, respectively. We can find that after a layer of GNN computation, the information of pairs-wise nodes transferes and aggregates with each other in the each homogeneous relation subgraph. Inspired by R-GCNs\cite{schlichtkrull2018modeling}, we devised the effective heterogenous aggregation method for node information from different pairs-wise subgraphs after messaging, As shown in Fig. 3. The nth layer heterogenous GNN is calculated as follows.
\begin{align}
	F^n_{Hetero}=ReLU(Aggr_{i \in \mathcal{E} }&(\sum_{v_i\in \mathcal{G}_{i}}^{}  (h^{(n-1)}(v_i)\cdot W^{(n)}_{1_i}+ \\ \notag
	&\sum_{w_i \in N(v_i)}h^{(n-1)}(w_i)\cdot W^{(n)}_{2_i}))) ,
\end{align}
where $F^n_{Hetero}$ refers to the operator of the nth layer of the heterogeneous GNN. $v_i$ refers to the node in the subgraph named i. 

It has been verified that GNN layers typically perform optimally with 2-3 layers. Therefore, this work use a two-layer heterogeneous GNN, as illustrated in Fig. 2. 
\subsection{Deep Reinforcement Learning for Crowd Navigation}
In most of the previous studies\cite{chen2019crowd}, \cite{yang2023st}, the objective of DRL is to determine the optimal policy that maximizes the state-action value function $Q^*(s_t,a_t)$:
\begin{equation}
	\pi^{*}(s_t)=\arg \max _{a_t}Q^*(s_t,a_t),
\end{equation}
then $Q^*(s_t,a_t)$ is established by the Bellman optimal equation:
\begin{align} 
	Q^*(s_t,a_t)= &\sum_{s_{t+1},r_t}P(s_{t+1},r_t \mid s_t,a_t) \notag \\
	&\times [r_t+\gamma \max_{a_{t+1}}Q^*(s_{t+1},a_{t+1})], 
\end{align}
where $\gamma \in (0,1)$ is a discount factor.

This letter introduces a discrete action space with feasible options for the agent. At each time step t, the agent chooses an action (denoted as $a_t$) consisting of a speed and a heading angle. We offer 5 distinct speeds, exponentially distributed between 0 and $v_{pref}$, and 16 evenly spaced heading angles across 0 to $2\pi$, resulting in a total of 80 fine-grained discrete actions.

As the scenario continues to expand, the previous setting of mismatched rewards makes the training less convergent\cite{chen2020relational}. Hence, this letter adopts the shaped reward proposed in SG-DQN\cite{zhou2022robot}. The reward $R^t$ with time step t includes rewards for reaching goals, comfort rewards, penalties for collision, etc. Please refer to SG-DQN for details. This optimized reward function propels agents to peak performance, bypassing the need for imitation learning initialization. We adhere primarily to algorithmic flow of the ST$^2$\cite{yang2023st} for anticipating a fairer contrast in subsequent comparisons.

We use  a 4-layer MLP with ReLU activations as a value estimator to approach $Q^*(s_t,a_t)$, As shown in Fig. 2.
\section{EXPERIMENTS}
\begin{figure}[t]
    \hspace{-4.5mm}
    \subfloat[] 
    {
        \includegraphics[width=0.245\textwidth]{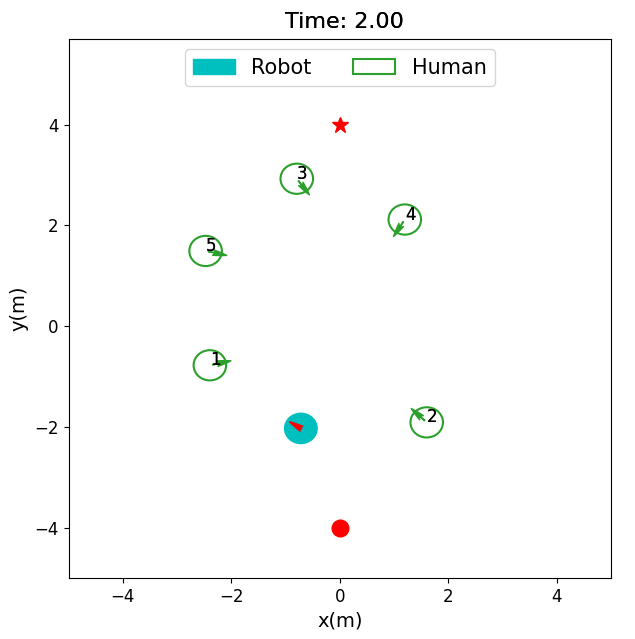}
    }\quad
    \hspace{-6mm}
    \subfloat[]
    {
        \includegraphics[width=0.245\textwidth]{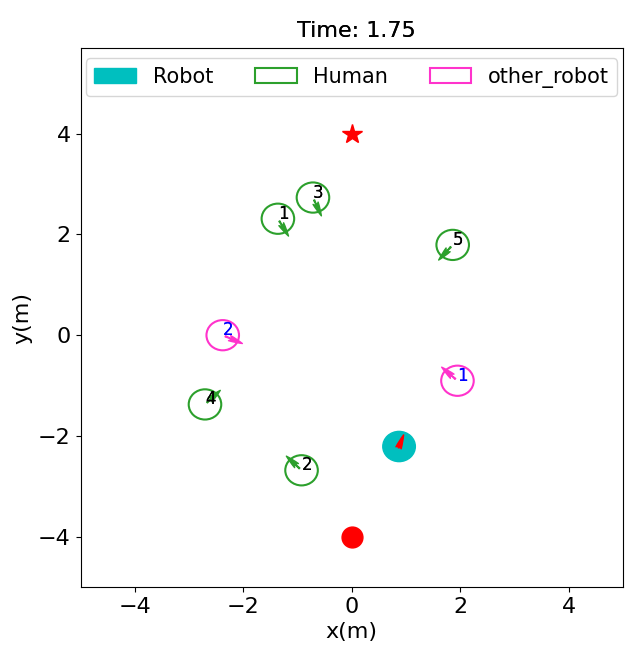}
    }
\caption{\textbf{Demonstration of our simulation environment.} We mainly utilize two training and testing scenarios: (a) Single-robot circle crossing scenario; (b) Multi-robot circle crossing scenario.}
\end{figure}
\begin{figure*}[t]
    \subfloat[LSTM-RL(Single-Robot)]
    {
        \includegraphics[width=0.18\textwidth]{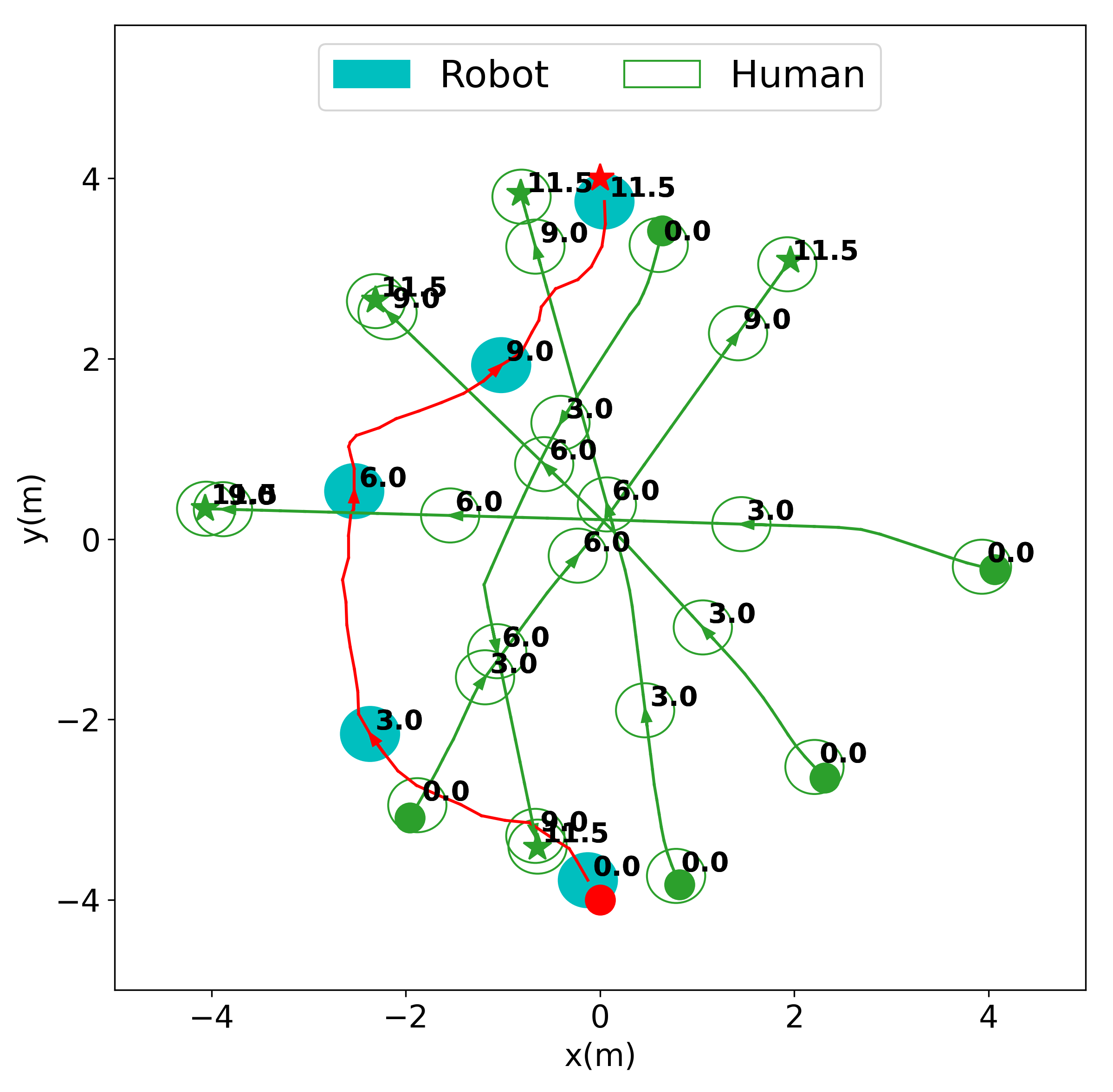}
    }
    \subfloat[LM-SARL(Single-Robot)]
    {
        \includegraphics[width=0.18\textwidth]{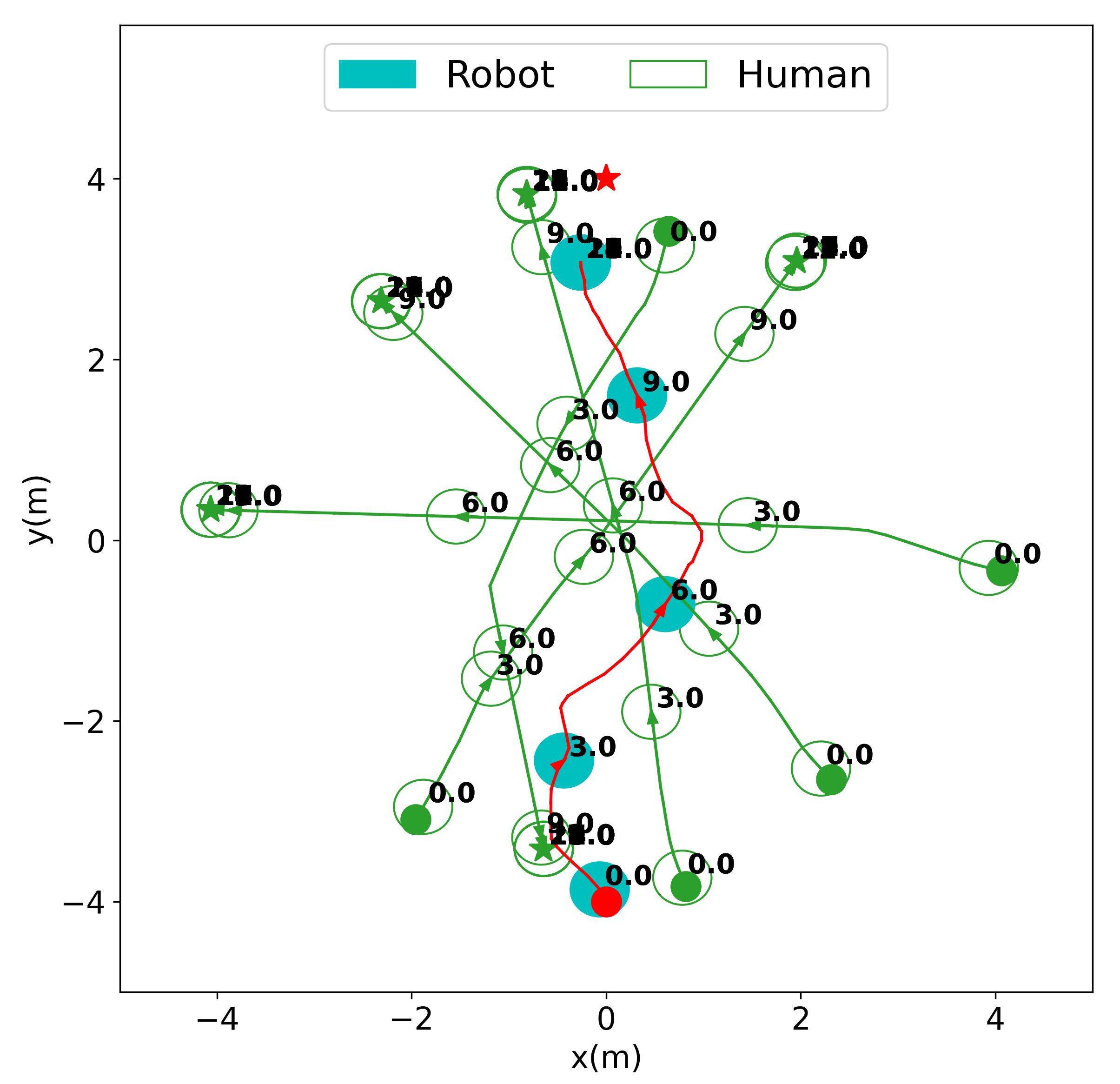}
    }
    \subfloat[ST$^2$(Single-Robot)]
    {
    	\includegraphics[width=0.18\textwidth]{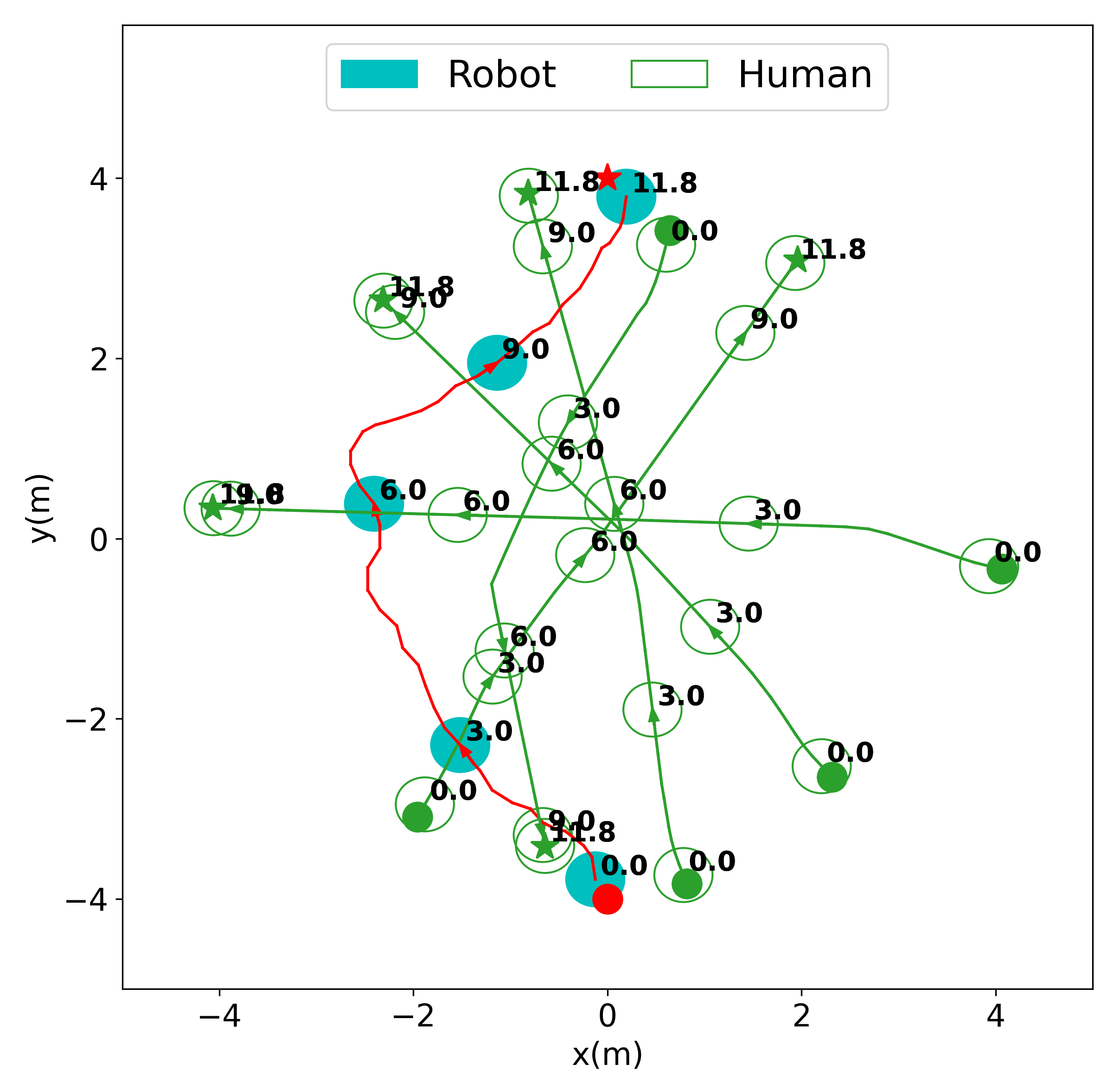}
    }
    \subfloat[HoR-DRL(Single-Robot)]
    {
    	\includegraphics[width=0.18\textwidth]{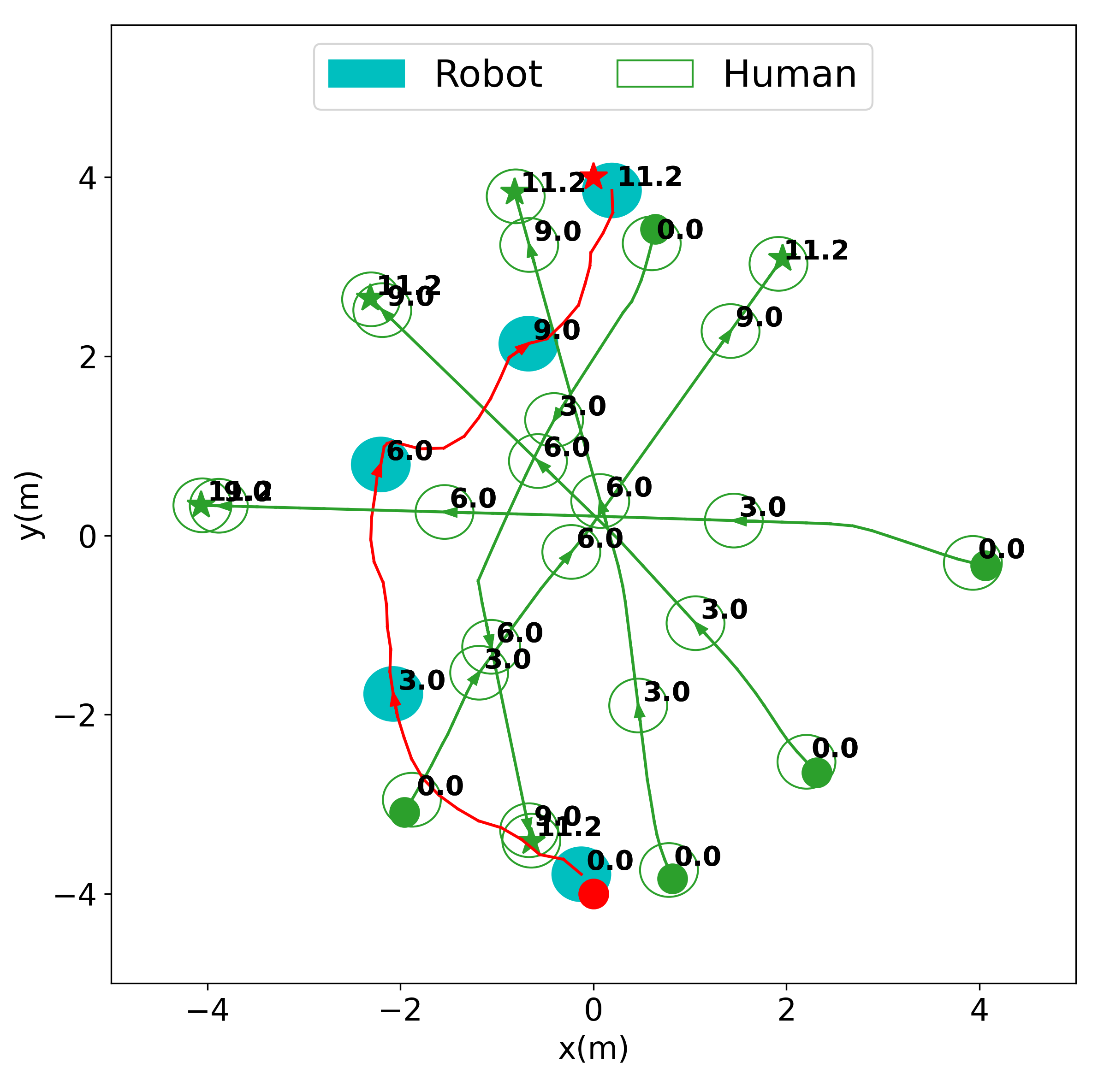}
    }
    \subfloat[HeR-DRL(Single-Robot)]
    {
        \includegraphics[width=0.18\textwidth]{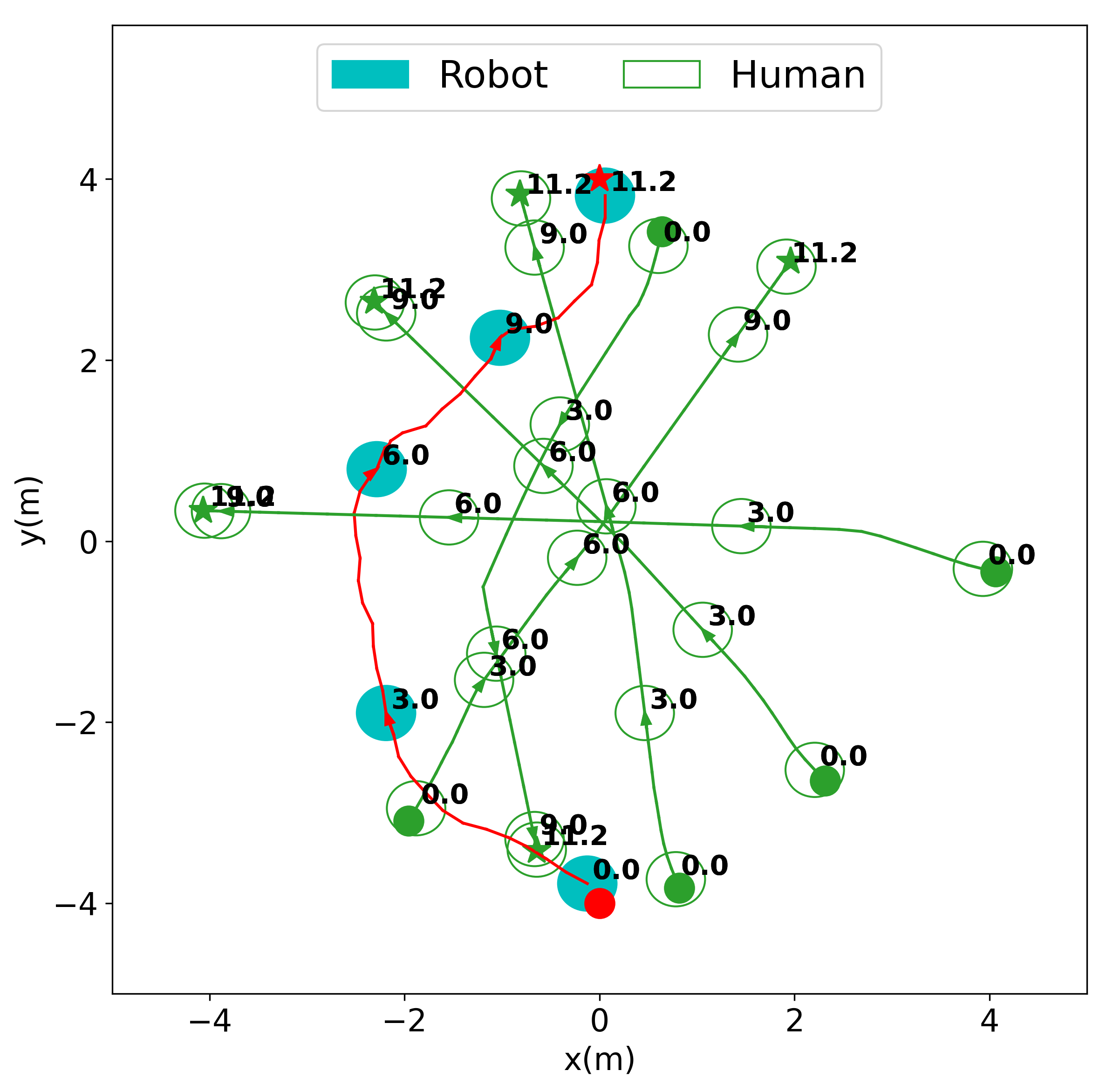}
    }
    
    \subfloat[LSTM-RL(Multi-Robot)]
    {
        \includegraphics[width=0.18\textwidth]{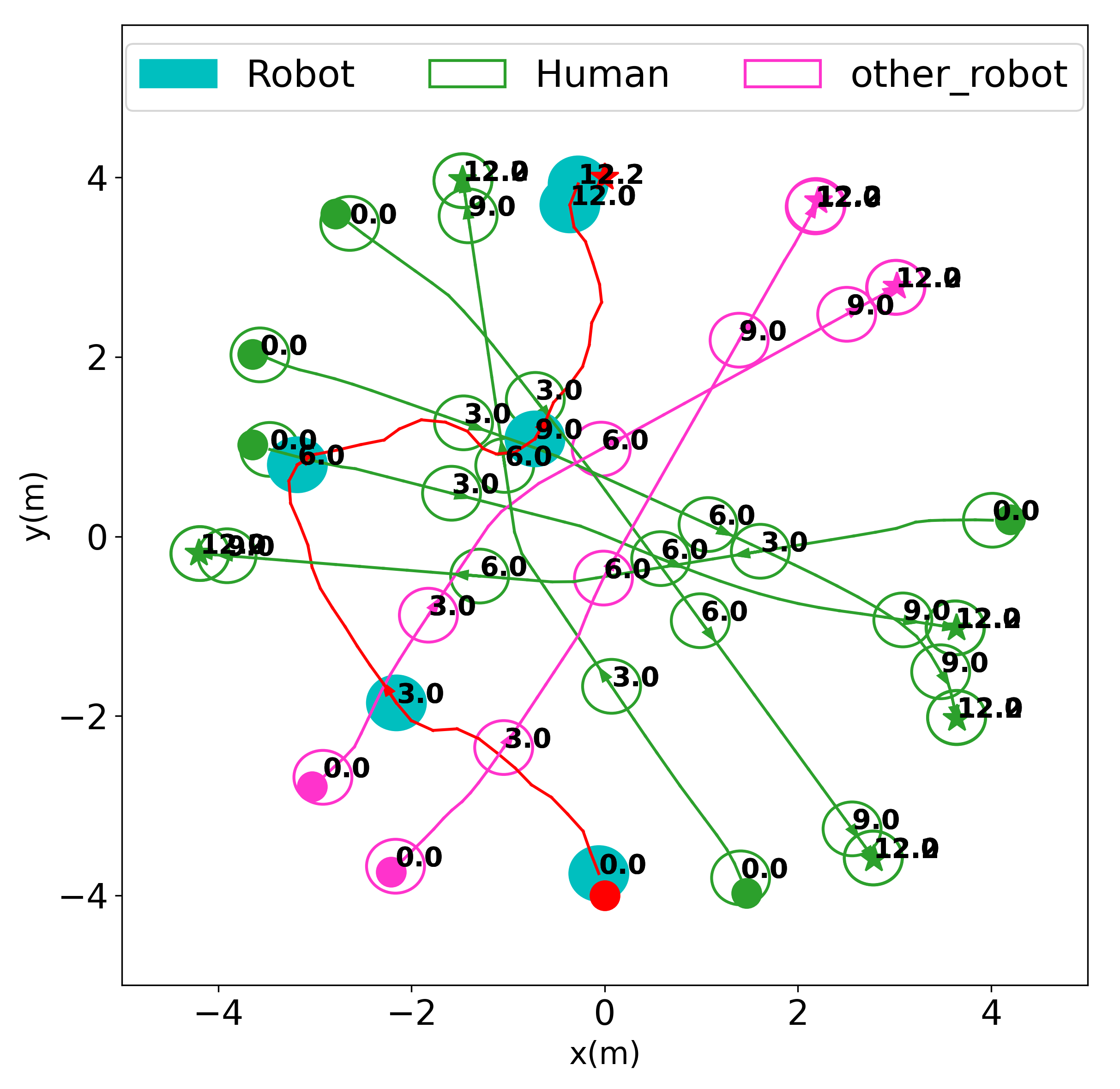}
    }
    \subfloat[LM-SARL(Multi-Robot)]
    {
        \includegraphics[width=0.18\textwidth]{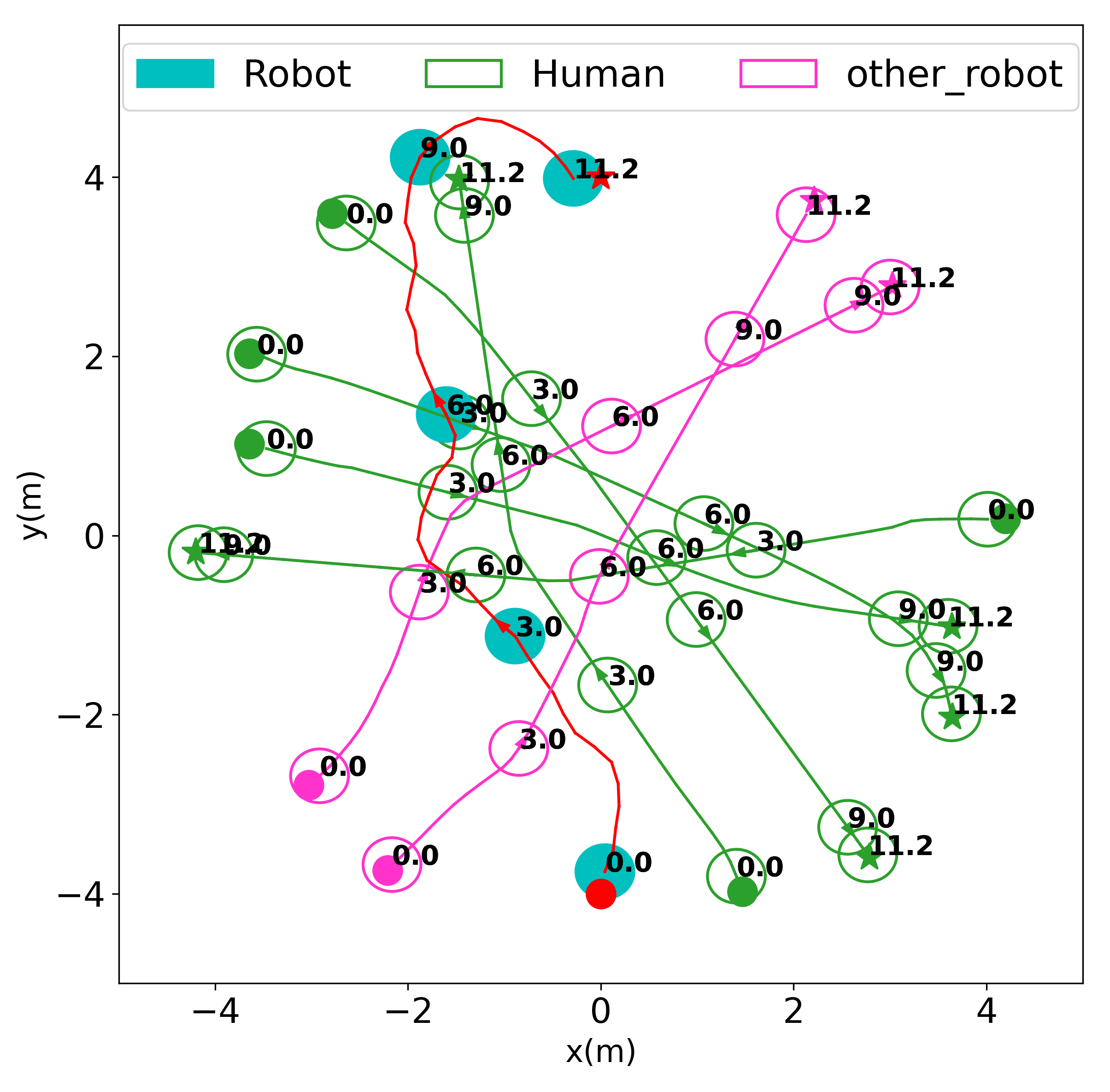}
    }
    \subfloat[ST$^2$(Multi-Robot)]
    {
        \includegraphics[width=0.18\textwidth]{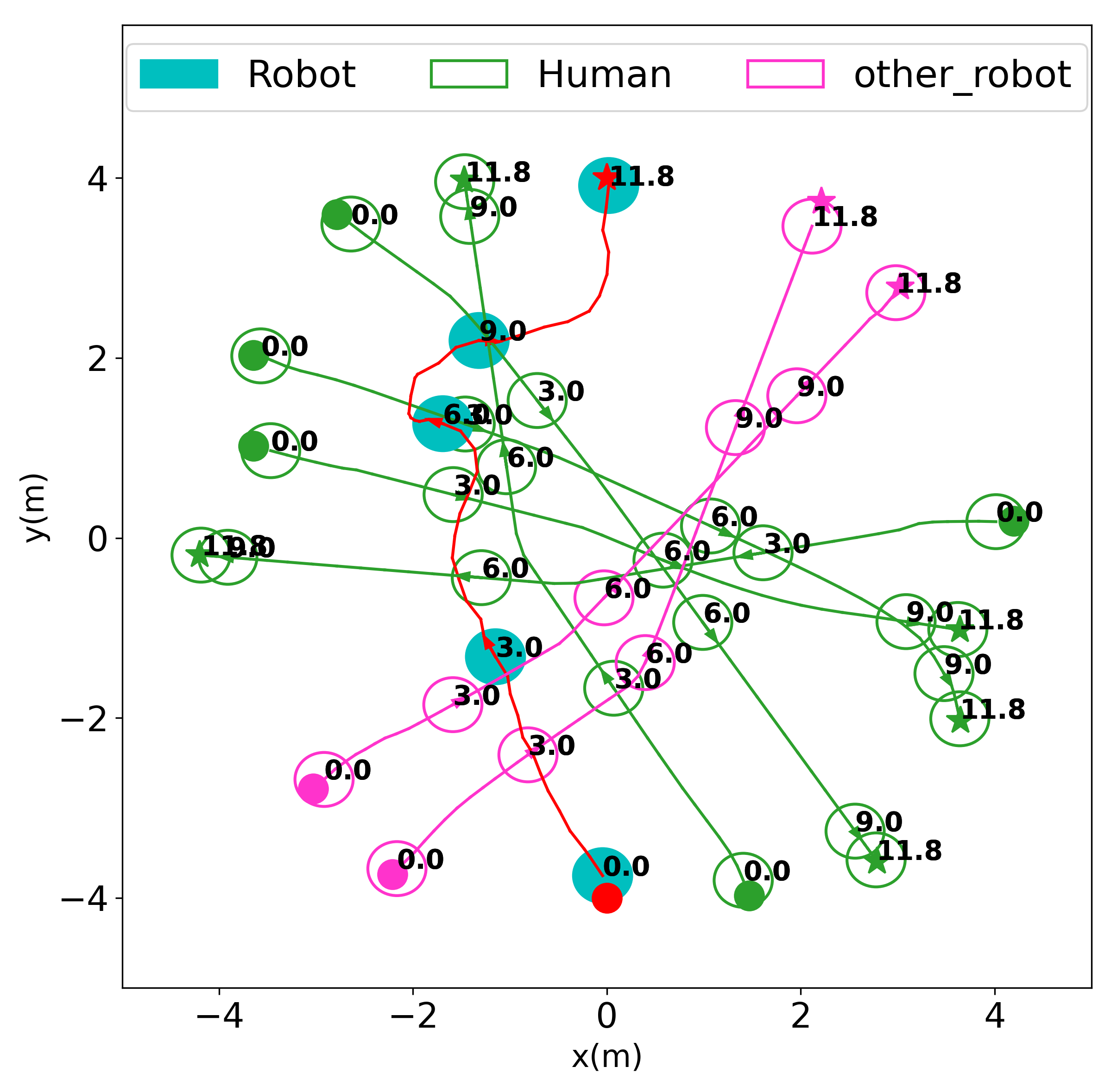}
    }
    \subfloat[HoR-DRL(Multi-Robot)]
    {
    	\includegraphics[width=0.18\textwidth]{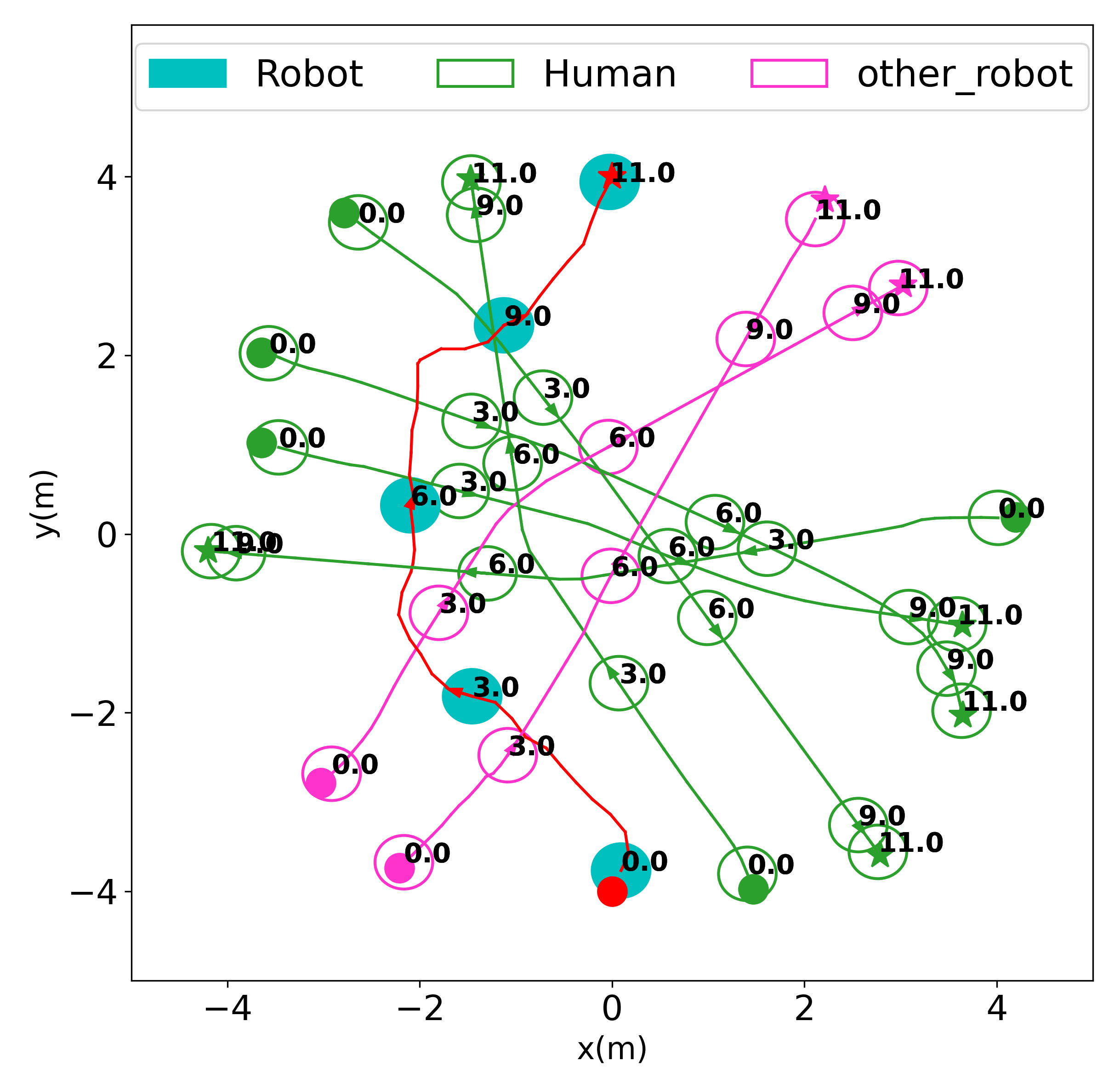}
    }
    \subfloat[HeR-DRL(Multi-Robot)]
    {
    	\includegraphics[width=0.18\textwidth]{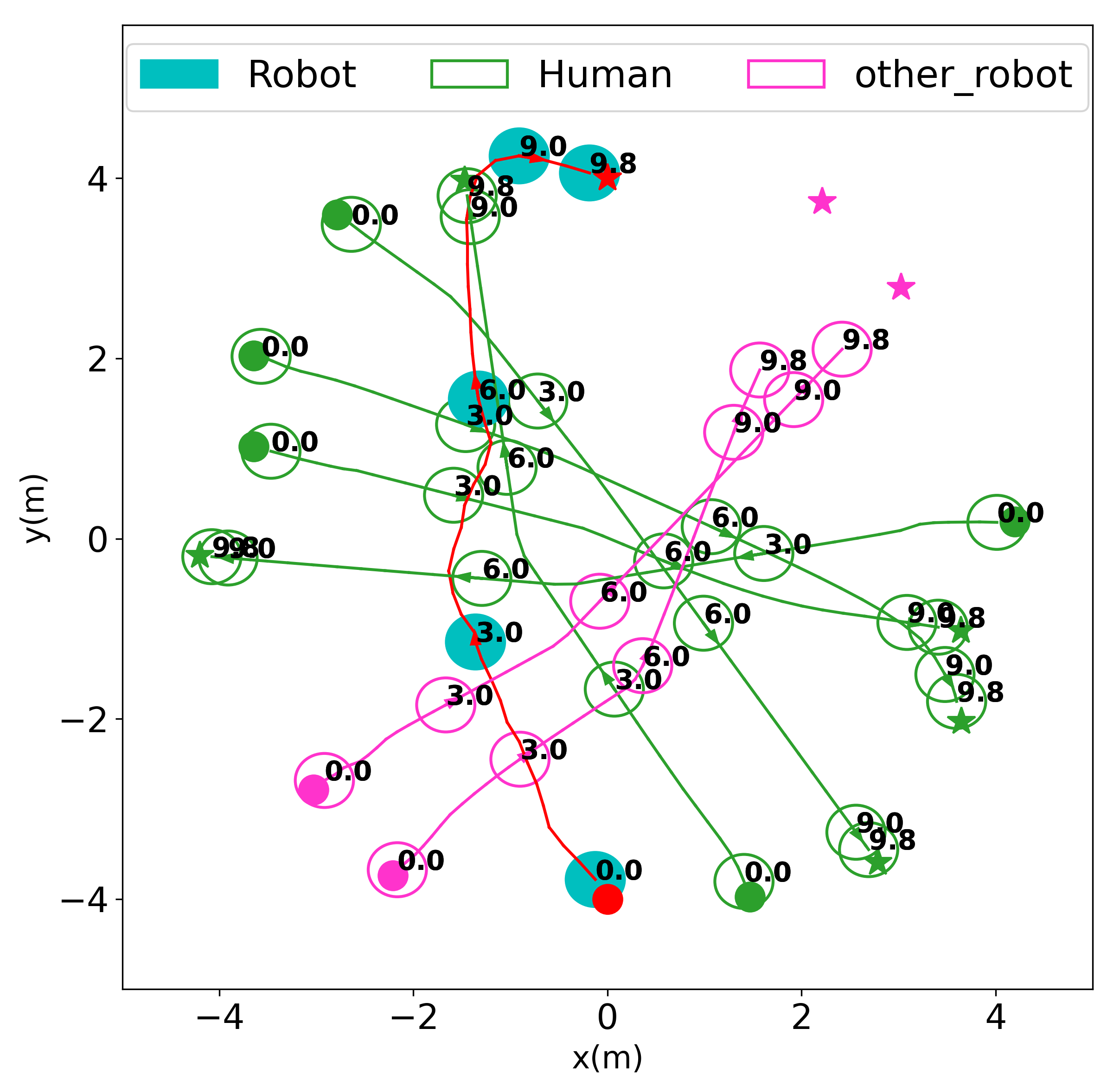}
    }
\caption{The comparison of navigation trajectories in single-robot and multi-robot circle crossing.}
\end{figure*}
\subsection{Simulation Setup}
\subsubsection{Simulation Environment}
Fig. 4 displays our 2D simulation environment based on CrowdNav\footnote{https://github.com/vita-epfl/CrowdNav.git}. Test scenarios are categorized into two groups: one is the single-robot circle crossing scenario, and the other is the multi-robot circle crossing scenario. The single-robot circle crossing scenario is set up with 5 humans, or 5H for short, while the multi-robot circle crossing scenario is set up with 5 humans and 2 other robots, or 5H2O for short. Unlike center robot and other robots, humans have comfort requirements and invisible settings to avoid training overly aggressive strategies. We employ unicycle kinematics for all agents. The human and other robots are controlled by ORCA. We also assume that all agents can achieve the desired velocities immediately, and they will keep moving with these velocities for $\delta t$ seconds.
\subsubsection{Baselines and Ablation Models}
We choose four existing state-of-the-art methods as baseline, including LSTM-RL\cite{everett2018motion} , LM-SARL\cite{chen2019crowd} and ST2\cite{yang2023st}. To assess the contributions of the heterogeneous GNN and the newly proposed parameters of the category on the performance, thus we set up the ablation models separately. Where HoR-DRL-nocate denotes the policy using a homogenous GNN without state information of the category. HoR-DRL denotes the policy using a homogenous GNN with state information of the category. HeR-DRL-nocate denotes the policy using the heterogeneous GNN without state information of the category. It is noteworthy that this homogeneous GNN particularly takes the states of all agents as nodes in a comprehensive graph, feeding them into the 2-layer GNN in Equation 4. Since there is only one class of agent in addition to the center robot in the single-robot scenario, only the HoR-DRL ablation method is trained. 
\subsubsection{Evaluation Metrics}
The performance metrics include \textit{Success Rate}, \textit{Collision Rate}, \textit{Average Time}, \textit{Discomfort Rate} and \textit{Min Distance}. \textit{Success Rate}(SR): the ratio of the center robot reaching its goal without a collision. \textit{Collision Rate}(CR): the ratio of the center robot colliding with other agents. \textit{Average Time}(AT): the average time taken by the center robot to reach its goal (in seconds). \textit{Discomfort Rate}(DR): the ratio of the minimum separating distance between a robot and other agents is less than risk but has not yet collided. \textit{Min Distance}(MD): the average robot minimum separating distance to the nearest obstacle agent in risk cases. Among them, SR and CR can reflect safety, AT reflects efficiency, and DR and MD reflect comfort.
\subsubsection{Training Process}
We implemented the policy in PyTorch\cite{paszke2017automatic} and trained it using the Adam\cite{kingma2014adam} optimizer with a learning rate of $1\mathit{e}^{-3}$. The batch size is set to 100 and discount factor $\gamma =0.9$. Due to differences in scenario complexity, for optimal performance after convergence, each method can be trained on 10,000 episodes for single-robot circle crossing  scenario and 15,000 episodes for multi-robot circle crossing scenario. The same parameter settings were used for training in both scenarios. Subsequent to each episode, the agent's position and goal point are randomly resampled, with the exception of the center robot. The exploration rate of the $\epsilon$-greedy policy decays linearly from 0.5 to 0.1 in the first $4k$ episodes and keeps 0.1 for the other episodes. To ensure fairness, deep reinforcement learning training process applied to all compared methods are uniform.

\subsection{Quantitative Evaluation}
\subsubsection{Performance Comparison with Baseline In Original Configuration}
Table I presents the test results of the original configuration in the single-robot scenario and the multi-robot scenario, respectively. HeR-DRL convincingly outperformed all baselines in both scenarios, with respect to SR, DR and MD. meanwhile, the advantage of HeR-DRL is larger in multi-robot scenarios than in single-robot scenarios. The analysis suggests that the superior performance in multi-robot scenarios is likely attributed to the presence of five heterogeneous relations, compared to the two(HHI and HCRI) in single-robot scenarios. This increased diversity of interactions enhances the effectiveness of the method. Analyzing the performance of AT, it is evident that our method exhibits a relatively longer overall navigation time. This can be attributed to the deliberate choice of paths prioritizing comfort. Opting for paths that enhance comfort may involve navigating through long detours or adopting lower speeds, ultimately leading to less efficient navigation. However, the difference in navigation time is not significant relative to the other methods, with a difference of 0.248s from the minimum in the single-robot scenario and 0.428s from the minimum in the multi-robot scenario. Meanwhile the frequency of discomfort improved significantly, with a minimum reduction of 34.2\% for single-robot scenarios and 33.3\% for multi-robot scenarios.
We were very pleased to see the positive results of sacrificing a small amount of time for a substantial increase in comfort.
\begin{table}[h]
	\caption{\centering QUANTITATIVE RESULTS IN CIRCLE CROSSING ENVIRONMENT}
	\label{table1}
	\centering
	\resizebox{\linewidth}{!}{
		\begin{tabular}{ccccccc} 
			\toprule
			Scenario&Method  &SR($\%$)&CR($\%$)&AT(s)&DR($\%$)&MD(m)\\
			\midrule
			\multirow{5}{*}{Single-Robot}&LSTM-RL      & 95.5 & 4.5     & 10.167 & 0.06 & 0.16 \\
			&LM-SARL         & 98.5 & 1.3     & 10.187 & 0.038 & $\textbf{0.17}$ \\
			&ST$^2$      	 & 98.9   & 1.1       & $\textbf{10.038}$  & 0.04 & $\textbf{0.17}$  \\
			&HoR-DRL  & 97.7 & 2.3    & 10.328  & $\textbf{0.022}$ & $\textbf{0.17}$  \\
			&HeR-DRL(Ours)  & $\textbf{99}$ & $\textbf{1}$    & 10.415  & 0.025 & 0.16  \\
			\midrule
			\multirow{7}{*}{Multi-Robot}&LSTM-RL      & 89.2 & 10.7      & 11.776  & 0.03 & 0.13\\
			&LM-SARL         & 94.1 & 4.7      & $\textbf{10.698}$  & 0.039 & 0.15\\
			&ST$^2$        & 95.5 & 4.5        & 10.949    & 0.035 & 0.14\\
			&HoR-DRL-nocate  & 93.3 & 6.6     & 10.841    & 0.025 & 0.15\\
			&HoR-DRL    & 95.7 & 3.6      & 10.779    & 0.021 & 0.15\\
			&HeR-DRL-nocate(Ours)  & 95.8 & 4.1      & 10.964  & 0.034 & 0.15  \\
			&HeR-DRL(Ours)  & $\textbf{96.3}$ & $\textbf{3.5}$      &  11.126  & $\textbf{0.02}$ & $\textbf{0.16}$  \\
			\bottomrule
		\end{tabular}
	}
\begin{tablenotes} 
\footnotesize
\item Bold values highlight the top performers within each environment configuration.
\end{tablenotes}
\end{table}
\subsubsection{Scenario Adaptation Comparison with Baseline}
In order to test the algorithm's scenario adaptation, we tested the trained algorithm under different configurations. To ensure fair and meaningful comparisons, we limited the scenario configurations to a maximum of 10 agents except center robot for mitigating potential performance crashes in overly dense environments. The selected test configurations include 5H3O, 5H4O, 5H5O, 6H2O, 7H2O, and 8H2O. The experimental results are presented in Table II. Table II unveils a clear trend: the overall performance of all algorithms consistently declines as the number of agents in the configuration increases. This inverse relationship is visually evident, suggesting a direct impact of agent density on algorithm effectiveness. Nevertheless, in terms of comfort, the HeR-DRL has a substantial advantage. Obviously, although HeR-DRL performs sub-optimally in terms of the SR and slightly inferior to ST$^2$, the CR, DR and MD metrics are overall superior to ST$^2$. Fortunately, the AT metrics of HeR-DRL lags behind relative to it by 0.77 s on average. This further reinforces the point mentioned in the previous section: the current methodology tends to trade a small amount of navigational efficiency for a significant improvement in comfort. Since only HRI was coded and no distinction was made between HCRI and HORI accordingly, It's apparent that LSTM-RL exhibits the worst overall performance. While LM-SARL demonstrates the best navigational efficiency, a substantial performance gap exists in other metrics when compared to ST$^2$ and HeR-DRL. This further underscores the importance of encoding complex heterogeneous relations.
\begin{table}[h]
	\caption{\centering QUANTITATIVE ADAPTATION ANALYSIS IN MULTIPLE CONFIGURATIONS OF MULTI-ROBOT CROWD NAVIGATION }
	\label{table2}
	\centering
	\resizebox{\linewidth}{!}{
		\begin{tabular}{ccccccc} 
			\toprule
			Methods  &Configure    &SR($\%$)&CR($\%$)&AT(s)&DR($\%$)&MD(m)\\
			\midrule
			\multirow{6}{*}{LSTM-RL}&5H3O         & 83.3 & 7.8    & 11.672 & 0.032 & 0.14 \\
			&5H4O      & 84.2 & 7.7     & 11.782 & 0.033 & 0.16 \\
			&5H5O         & 82.5 & 9.1     & 12.079 & 0.032 & 0.15 \\
			&6H2O       	 & 80.4   & 8.1       & 11.734  & 0.047 & 0.14  \\
			&7H2O  & 75.8 & 14.3    & 12.187  & $\textbf{0.045}$ & 0.13  \\
			&8H2O  & 68.5 & 18.1    & 12.411  & 0.06 & 0.14  \\
			\midrule
			\multirow{6}{*}{LM-SARL}&5H3O         & 88.3 & 10.1    & $\textbf{10.7}$ & 0.044 & 0.14 \\
			&5H4O      & 85.8 & 13.3     & $\textbf{10.794}$ & 0.052 & 0.14 \\
			&5H5O         & 83.9 & 14.7     & 10.942& 0.062 & 0.13 \\
			&6H2O       	 & 87.7   & 11.2       & $\textbf{10.859}$   & 0.057 & 0.14  \\
			&7H2O  & 84.6 & 14.5    & $\textbf{11.011}$  & 0.079 & 0.14  \\
			&8H2O  & 79.6 & 18.5    & $\textbf{11.196}$  & 0.112 & 0.14  \\
			\midrule
			\multirow{6}{*}{ST$^2$}&5H3O         & $\textbf{95.7}$ & $\textbf{4}$    & 10.887 & 0.037 & 0.15 \\
			&5H4O      & $\textbf{93.4}$ & 6.3     & 10.929 & 0.046 & 0.15 \\
			&5H5O         & $\textbf{92.7}$ & 7.2     & $\textbf{10.926}$ & 0.048 & 0.15 \\
			&6H2O       	 & $\textbf{95.2}$  & $\textbf{4.4}$    & 11.087  & 0.046 & 0.14  \\
			&7H2O  & $\textbf{91.4}$ & 8.2    & 11.139  & 0.063 & 0.15  \\
			&8H2O  & $\textbf{89}$ & $\textbf{10.5}$    & 11.235  & 0.082 & $\textbf{0.15}$  \\
			\midrule
			\multirow{6}{*}{HeR-DRL}&5H3O         & 93.4 & 4.3    & 11.283 & $\textbf{0.021}$ & $\textbf{0.17}$ \\
			&5H4O      & 91.5 & $\textbf{4.8}$     & 11.653 & $\textbf{0.025}$ & $\textbf{0.17}$ \\
			&5H5O         & 85.6 & $\textbf{6.3}$      & 12.492 & $\textbf{0.02}$ & $\textbf{0.17}$ \\
			&6H2O       	 & 92.3   & 4.9       & 11.434  & $\textbf{0.035}$ & $\textbf{0.17}$  \\
			&7H2O  & 87.5 & $\textbf{7.9}$    & 11.756  & $\textbf{0.045}$ & $\textbf{0.16}$  \\
			&8H2O  & 80.6 & 11.3    & 12.216  & $\textbf{0.059}$   & $\textbf{0.15}$  \\
			\bottomrule
		\end{tabular}
	}
\begin{tablenotes} 
\footnotesize
\item Bold values highlight the top performers within each environment configuration.
\end{tablenotes}
\end{table}
\subsection{Qualitative Evaluation}
We further investigated the effectiveness and superiority of our method by performing trajectory analyses in single-robot scenarios and multi-robot scenarios respectively, as shown in Fig.5. Firstly, in the test case of the single robot scenario, LM-SARL crosses the pedestrians from the middle and freezes at the end, and LSTM-RL, ST$^2$, and HeR-DRL all choose to go around the middle to reach the end point. However, it can be clearly seen that HeR-DRL's trajectory is the one with the least abrupt change in the direction. Then in the multi-robot circle crowssing scenario. LSTM-RL and ST$^2$ have similar trajectories, favouring obstacle avoidance and bypassing from after the agent, while LM-SARL and HeR-DRL avoid obstacles and bypass from before the agent. After comparison, LSTM-RL deliberately avoids from the back of the trajectory after the 6th second, resulting in a long route around the far side. ST$^2$ performs better than LSTM-RL in terms of smoothness, but its average distance from the other agents during the movement is significantly worse than that of HeR-DRL. Although the trajectories of LM-SARL and HeR-DRL are similar, the trajectory show that LM-SARL tends to go around at a low speed even though it can accelerate in front of it, while HeR-DRL accelerates when it has the conditions to accelerate, which gives it an advantage in terms of direction change and efficiency. Despite focusing only on spatial aspects, HeR-DRL achieves comfort and trajectory quality equal to or even better than ST2, which models both spatial and temporal interactions.
\subsection{Ablation Analysis}
Comparing HoR-DRL-nocate and HoR-DRL reveals that the addition of state information of the category not only enhances the success rate but also improves the comfort metrics. Similar conclusions can be drawn from the performance of the comparison between HeR-DRL-nocate and HeR-DRL. This indicates that the state information of the category comprehensively improves the extraction of interactive relations in scenarios.

Comparison of the two ablation models, HoR-DRL-nocate and HeR-DRL-nocate, revealed that HeR-DRL-nocate sacrificed a certain level of efficiency and comfort to improve the success rate. The same conclusion was found when comparing the HoR-DRL and HeR-DRL in single-robot scenario. This seems to imply that our proposed heterogeneous GNN are more inclined to improve success rates, even at the expense of a small amount of comfort. Nevertheless, a comparison of HoR-DRL and HeR-DRL in a multi-robot scenario reveals that the addition of the heterogeneous GNN leads to an improvement in overall performance in addition to AT. The analysis reveals that incorporating state information with the category enhances the effectiveness of the heterogeneous GNN in managing diverse relations. This enables a notable increase in the success rate while maintaining, or possibly slightly improving, overall comfort. Trajectory analysis of HoR-DRL and HeR-DRL in both scenarios demonstrates that HeR-DRL has fewer changes in the direction of motion than HoR-DRL, and the trajectory is smoother, reflecting the excellence of heterogeneous GNN for modelling interactive relations.
\section{CONCLUSION}
In this letter, HeR-DRL is proposed to solve the navigation problem in multi-robot pedestrian scenarios. The proposed algorithm fully improves the representation of scenario interactive relations by constructing a robot-crowd heterogeneous relation graph with the help of the state information of the category. The experimental results show that the proposed method achieves the best performance in terms of success rate and comfort metrics, in contrast to the state-of-the-art baselines.
However, there are some limitations to the methodology of this letter, which only explores spatial heterogeneity and does not address the effects of spatial-temporal heterogeneity. Thus, This will be explored in future research.

\small
\bibliographystyle{ieeetr}

\end{document}